\documentclass[sigconf, screen, nonacm]{acmart}

\usepackage[]{hyperref}

\usepackage{multirow}
\usepackage{tikz}
\usepackage{amsmath}
\usepackage{xspace}
\usepackage{natbib}
\usepackage{enumitem}
\usepackage[ruled,vlined, linesnumbered, noend]{algorithm2e}
\usepackage{algorithmic} 
\usepackage{pifont}
\usepackage{xfrac}
\usepackage{balance}

\begin{document}
\sloppy{
\emergencystretch 3em
}

\newcommand{\iid}{\texttt{iid}\xspace}
\newcommand{\noniid}{non-\texttt{iid}\xspace}
\newcommand{\systemName}{\texttt{EcoLearn}\xspace}
\newcommand{\cselect}{\texttt{CSelect}\xspace}
\newcommand{\cprov}{\texttt{CProv}\xspace}
\newcommand{\cscale}{\texttt{CScale}\xspace}
\newcommand{\emissionunit}{\ensuremath{g\cdot CO_{2}eq}\xspace}
\newcommand{\ciunit}{\ensuremath{g\cdot CO_{2}eq/kWh}\xspace}
\newcommand{\cciunit}{\ensuremath{g\cdot CO_{2}eq/cycle}\xspace}
\newcommand{\efunit}{\ensuremath{cycle/kWh}\xspace}
\newcommand{\xmark}{\ding{55}}%
\newcommand{\tickmark}{\ding{51}}%

\newcommand{\noman}[1]{  
	{\textcolor{red}{(\textbf{Noman says:}  #1)}}{}}

 \newcommand{\talha}[1]{  
	{\textcolor{blue}{(\textbf{Talha:}  #1)}}{}}
 
\newcommand{\remove}[1]{  
	{\textcolor{red}{(\textbf{Potentially remove:}  #1)}}{}}

\date{}



\title[EcoLearn]{EcoLearn: Optimizing the Carbon Footprint\\of Federated Learning}

\author{Talha Mehboob}
\affiliation{
  \institution{University of Massachusetts Amherst}
    \city{}
  \country{}
}
\email{tmehboob@umass.edu}

\author{Noman Bashir}
\affiliation{
  \institution{Massachusetts Institute of Technology}
    \city{}
  \country{}
}
\email{nbashir@mit.edu}

\author{Jesus Omana Iglesias}
\affiliation{
  \institution{Telefonica Research}
    \city{}
  \country{}
}
\email{jesusalberto.omana@telefonica.com}

\author{Michael Zink}
\affiliation{
  \institution{University of Massachusetts Amherst}
    \city{}
  \country{}
}
\email{mzink@cas.umass.edu}

\author{David Irwin}
\affiliation{
  \institution{University of Massachusetts Amherst}
    \city{}
  \country{}
}
\email{irwin@ecs.umass.edu}

\begin{abstract}
Federated Learning (FL) distributes machine learning (ML) training across edge devices to reduce data transfer overhead and protect data privacy. Since FL model training may span hundreds of devices and is thus resource- and energy-intensive, it has a significant carbon footprint.  Importantly, since energy's carbon-intensity differs substantially (by up to 60$\times$) across locations, training on the same device using the same amount of energy, but at different locations, can incur widely different carbon emissions. While prior work has focused on improving FL's resource- and energy-efficiency by optimizing time-to-accuracy, it implicitly assumes all energy has the same carbon intensity and thus does not optimize carbon efficiency, i.e., work done per unit of carbon emitted.   

To address the problem, we design \systemName, which minimizes FL's carbon footprint without significantly affecting model accuracy or training time. \systemName achieves a favorable tradeoff by integrating carbon awareness into multiple aspects of FL training, including i) selecting clients with high data utility and low carbon, ii) provisioning more clients during the initial training rounds, and iii) mitigating stragglers by dynamically adjusting client over-provisioning based on carbon. We implement \systemName and its carbon-aware FL training policies in the Flower framework and show that it reduces the carbon footprint of training (by up to $10.8$$\times$) while maintaining model accuracy and training time (within $\sim$$1$\%) compared to state-of-the-art approaches. 
\end{abstract}

\maketitle 

\section{Introduction}
\label{sec:intro}
Federated Learning (FL) is an increasingly popular machine learning (ML) approach that trains models on data distributed across numerous edge devices~\cite{bian2024cafe, federated-learning-paper}, potentially spread over large geographic areas. FL enhances data privacy and security compared to centralized ML by processing data locally and only sending updated model parameters to a central server. This reduces data transfer overhead and keeps raw data on the device, safeguarding user privacy. In many application scenarios, FL is an important tool for complying with data protection laws, such as GDPR~\cite{GDPR2016a}.

Unlike centralized learning, FL operates on distributed data across many clients. In each training round, only a subset of clients is selected, often randomly. These clients perform multiple training iterations, e.g., using stochastic gradient descent on their local data, before sending model updates to a central server for aggregation into the global model. FL models are typically trained on privacy-sensitive, heterogeneous data from diverse environments. For example, images from different CCTV cameras in hospital parking lots across the United States during winter may vary significantly, meaning data across clients may not be independent and identically distributed (i.i.d.).  Importantly, the clients selected in each round can impact i) the time required to reach a target accuracy or ii) the accuracy achieved within a fixed training time. For example, repeatedly selecting clients missing certain data classes may prolong training, requiring additional rounds to cover those classes. 

FL often selects random subsets of clients each round. However, this approach can increase training time by requiring many rounds to ensure sufficient data coverage.  Random selection may also waste resources by continually training on data that contributes little to model accuracy. Recent research focuses on intelligent participant selection, which chooses clients based on their data's statistical ``utility'' -- how much data improves the global model's accuracy~\cite{oort-osdi,refl-eurosys}.  Such intelligent client selection significantly reduces resource waste and shortens training time. Unfortunately, while reducing the training time necessary to achieve a target accuracy is important, prior work generally does not consider that clients may incur widely different ``costs'' for local training. In particular, an increasingly important cost is energy's carbon intensity, which may differ widely across geographically distributed clients due to varying sources of electric energy (e.g., fossil fuels, renewables, nuclear, etc.) in different locations. Thus, optimizing a model's carbon footprint is becoming just as important as optimizing its accuracy and training time. 
Furthermore, recent work~\cite{sani2024future} suggests that by 2026~\cite{villalobos2024will}, advancements in foundational models will increasingly rely on FL, as organizations with computational resources but restrictive data policies opt for collaborative training~\cite{douillard2023diloco, sani2024photon, strati2024ml}. This shift has significant sustainability implications, a recent study~\cite{MLSYS2022} highlights that training a small ML task using FL generates higher carbon emissions than training large transformer models in a centralized setting. These challenges emphasize the need for research to enhance efficiency and sustainability in FL.

\begin{figure*}[t]
    \centering
    \includegraphics[width=1\linewidth]{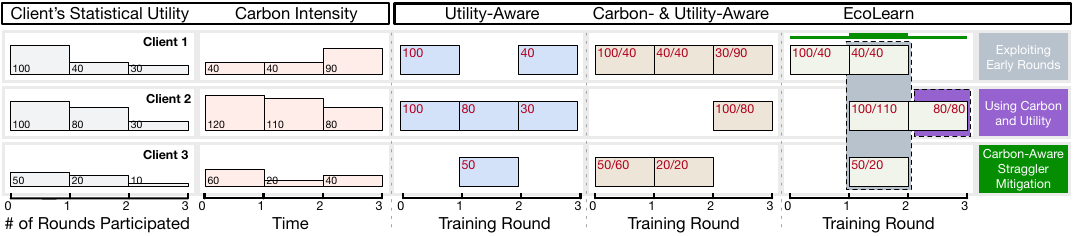}
    \vspace{-0.6cm}
    \caption{\textbf{\emph{Grey box highlights the selection of more clients in the initial rounds. Purple box shows that we use our basic utility and carbon-aware variant in the EcoLearn.}}}
    \label{fig:illustrative-example}
   \vspace{-0.45cm}
\end{figure*}

To address the problem, we present \systemName, which concurrently optimizes FL's carbon footprint, accuracy, and training time.  
As we discuss, gracefully navigating this three-way tradeoff requires integrating carbon awareness into multiple aspects of FL training, such as deciding which clients to select, how many to select, and when.  
As shown in~\autoref{fig:illustrative-example}, existing utility-based (and carbon-agnostic) client selection achieves a high model accuracy and low training time but incurs a high carbon cost~\cite{oort-osdi, refl-eurosys}.
Alternatively, a carbon-only policy that selects only low-carbon clients significantly increases training time and decreases accuracy due to insufficient data coverage (not shown). 

\systemName optimizes its carbon footprint while maintaining training time and accuracy by integrating carbon-awareness into \emph{when}, \emph{how many}, and \emph{which} clients to select. \systemName exploits the ``neuroplasticity'' of a model in early rounds, i.e., its \emph{critical learning period}~\cite{clp_accordion,clp_fl}, to bootstrap model weights using a large number of high-value clients. \systemName improves accuracy beyond a utility-based approach by selecting only high-utility clients during this early-round scale-up. 
However, it selects clients with high normalized utility per unit carbon, sacrificing the potential increase in accuracy beyond the utility-based approach to lower its carbon footprint, as shown in~\autoref{fig:illustrative-example}. 
The final ingredient is dynamic over-provisioning of clients for straggler mitigation, selecting more ``extra'' clients during low carbon periods to reduce training time, instead of a fixed ratio used by prior work~\cite{oort-osdi}. \systemName's techniques can generalize to ``cost'' metrics other than carbon.

Our hypothesis is that \systemName's adaptive carbon-aware policies for which, when, and how many clients to select can substantially lower FL's carbon footprint (by an order of magnitude) while maintaining model accuracy and training time, even for highly non-i.i.d. data, compared to state-of-the-art carbon-agnostic approaches.
In evaluating our hypothesis, we make the following contributions.

\begin{enumerate}[leftmargin=0.6cm, topsep=0.25cm, itemsep=0.1cm]
    \item {\bf Carbon Awareness in FL.} Optimizing FL's carbon footprint by leveraging spatiotemporal variations in the grid's carbon intensity presents a fundamental tradeoff between carbon, model accuracy, and training time. A singular focus on selecting \emph{which} clients for training cannot gracefully navigate this tradeoff.  To do so, we integrate carbon awareness into multiple aspects of client selection, as detailed in \autoref{sec:background}.

    \item \noindent {\bf Practical Adaptive Carbon-Aware Policies.} We design adaptive carbon-aware policies for \systemName that solve the challenges of precisely determining which, how many, and when to select clients when training an FL model in practice. \systemName's policies lie at the Pareto-frontier of model accuracy and carbon cost to train the model. 
    We provide the details of our client selection policies in \autoref{sec:design}.

    \item \noindent {\bf Implementation and Evaluation.} We implement \systemName using the Flower framework~\cite{flower}, and evaluate it across multiple synthetic and real datasets with different data distributions across clients. Our results show that \systemName reduces the carbon footprint of FL training (by up to $10.8$$\times$) while maintaining model accuracy and training time of state-of-the-art approaches. \autoref{sec:evaluation} presents the detailed evaluation.

\end{enumerate}

\section{Background and Motivation}
\label{sec:background}
Below, we provide an overview of both FL (\S\ref{sec:fl-overview}) and prior work on resource- and carbon-efficient FL (\S\ref{sec:resource-efficient-fl}).  We then motivate \systemName's approach to carbon-efficient FL (\S\ref{sec:motivation}).  

\setlength{\tabcolsep}{2.0pt}
\RestyleAlgo{ruled}
\newlength{\textfloatsepsave} 
\setlength{\textfloatsepsave}{\textfloatsep} 
\setlength{\textfloatsep}{6pt} 
\begin{table}[t]
\small
\centering
\caption{List of variables and their descriptions.}
\vspace{-0.3cm}
\begin{tabular}{|c|l|}
\hline
\textbf{Variable}      & \textbf{Description}                                                                                         \\ \hline
$ \mathcal{K}_t $      & Set of clients selected to participate in round $t$.                                                         \\ \hline
$ N $                  & Total number of clients in the FL system.                                                                    \\ \hline
$ \mathbf{w}_t $       & Global model parameters at round $t$.                                                                        \\ \hline
$ D_k $                & Local dataset of client $k$.                                                                                 \\ \hline
$ \mathcal{L}_k(\mathbf{w}) $ & Local loss function for client $k$, based on dataset $D_k$.                                             \\ \hline
$ \ell(\mathbf{w}; x_i, y_i) $ & Loss function for a single data point $(x_i, y_i)$.                                                    \\ \hline
$ \Delta \mathbf{w}_k $& Model update (gradient) from client $k$.                                                                     \\ \hline
$ \eta $               & Learning rate for gradient descent.                                                                          \\ \hline
$ |D_{\text{total}}| $ & Total data points across all participating clients in a round.                                      \\ \hline
\end{tabular}
\label{tab:variables}
\end{table}

\vspace{-0.1cm}
\subsection{Federated Learning Overview}
\label{sec:fl-overview}
Federated Learning (FL) is a machine learning (ML) approach for iteratively training a model where the training data is distributed across many client devices~\cite{federated-learning-paper}. 
An FL framework comprises two main components: a centralized server or \emph{aggregator} and distributed devices or \emph{clients}. 
The training process occurs over multiple \emph{rounds} to improve model accuracy by leveraging numerous clients~\cite{fedavg-mlsys}.

In each round \( t \), the aggregator determines \emph{which} and \emph{how many} clients will participate.
Let \( \mathcal{K}_t \) be the set of clients in round \( t \), and \( N \) be the total number of clients. Typically, \( |\mathcal{K}_t| \ll N \) since only a small fraction of clients participate in each round due to communication bottlenecks.
Each participating client \( k \in \mathcal{K}_t \) receives the global model parameters \( \mathbf{w}_t \) and trains on its local dataset \( D_k \). The objective at each client is to minimize the local loss function \( \mathcal{L}_k(\mathbf{w}) \) on the local data \( D_k \):
\begin{equation*}
    \mathcal{L}_k(\mathbf{w}) = \frac{1}{|D_k|} \sum_{i \in D_k} \ell(\mathbf{w}; x_i, y_i)
\end{equation*}
where \( \ell(\mathbf{w}; x_i, y_i) \) represents the loss for a single data point \( (x_i, y_i) \), and \( \mathbf{w} \) denotes the model parameters. 

The clients compute the gradient update \( \Delta \mathbf{w}_k = \mathbf{w}_t - \eta \nabla \mathcal{L}_k(\mathbf{w}_t) \) using their local data. Each client sends the model update back to the server, which aggregates these updates. A common aggregation strategy is \emph{Federated Averaging (FedAvg)}~\cite{federated-learning-paper}, which updates the global model as a weighted average of the clients' updates:
\[
\mathbf{w}_{t+1} = \frac{1}{|\mathcal{K}_t|} \sum_{k \in \mathcal{K}_t} \frac{|D_k|}{|D_{\text{total}}|} \Delta \mathbf{w}_k
\]

where \( |D_{\text{total}}| \) is the total number of data points for all participating clients in that round.
However, some client updates may not arrive due to network issues, computational resource availability, or energy constraints. Clients that do not respond or take too long are called \emph{stragglers}, and can significantly delay training. To mitigate these stragglers, the aggregator may use techniques such as discarding updates from late-arriving clients.

FL typically includes criteria that determine when to stop training~\cite{dinh2020federated,federated-learning-paper}. We focus on two widely-used criteria: \emph{model convergence}, where training stops when gains in model performance become small, i.e., \( ||\mathbf{w}_{t+1} - \mathbf{w}_t||_2 < \tau \), where \( \tau \) is the threshold, and \emph{performance threshold}, where training halts when the model's accuracy on a validation dataset reaches a predefined threshold \( T_{\text{perf}} \).

\begin{table}[t]
    \centering
    \small
   \caption{\textbf{\emph{Summary of related FL frameworks and their key attributes compared to} \systemName.}}
   \vspace{-0.25cm}
        \begin{tabular}{|c|c|c|c|c|c|}
            \hline
            \textbf{FL} & \textbf{Client} & \textbf{Straggler} & \textbf{Client} & \textbf{Carbon}\\
            \textbf{Framework} & \textbf{Selection} & \textbf{Mitigation} & \textbf{Scaling} & \textbf{Awareness}\\ \hline \hline
             \textbf{Oort} & \tickmark & \tickmark & \xmark & \xmark \\ \hline
             \textbf{REFL} & \tickmark & \tickmark & \xmark & \xmark  \\ \hline
             \textbf{CriticalFL} & \xmark & \xmark & \tickmark  & \xmark \\ \hline
             \textbf{FedZero} & \tickmark & \tickmark & \xmark & \tickmark \\ \hline
             \textbf{PyramidFL} & \tickmark & \tickmark & \xmark & \xmark \\ \hline \hline
             \systemName & \tickmark & \tickmark & \tickmark  & \tickmark \ \\ \hline
        \end{tabular}
       \label{table:baseline-algs}
\end{table}

\subsection{Carbon-Aware and Resource-Efficient FL}
\label{sec:resource-efficient-fl}
In this section, we review work on improving FL's resource efficiency.
We present an analysis of spatiotemporal variations in energy's carbon intensity across clients. 
We also discuss how prior work on resource-efficient FL does not apply to optimizing carbon.

\noindent
\textbf{1 -- Resource efficiency in FL.} FL's resource efficiency is dictated by the operation of its aggregator and clients. 
Prior work has explored optimizing resource requirements for aggregators, e.g., \cite{googledatacenters}. 
However, clients incur most of the computational load in FL using their computational and communication resources. 
Prior work has advanced optimizing model performance~\cite{li2020fair} and system efficiency~\cite{federated-learning-paper} to tackle the challenges posed by clients' data and system heterogeneity. 
However, further optimizations, such as \textit{intelligent} client selection instead of the traditional random selection~\cite{fedavg-mlsys, MLSYS2022_a8bc4cb1}, have only recently received attention.

Recent work has improved on random selection. Oort~\cite{oort-osdi} employs a guided client selection approach to prioritize clients with higher statistical utility to maximize system efficiency. Statistical utility is measured using training loss as a proxy, while system efficiency is based on completion time. 
Oort favors faster clients to reduce round duration and uses a pacer algorithm that allows longer round durations that include unexplored or slower clients to improve overall statistical efficiency.
REFL~\cite{refl-eurosys} extends Oort to minimize resource waste by incorporating model updates from straggler clients as they arrive after each round ends. 
Other approaches, such as PyramidFL~\cite{pyramidFL-mobicom}, fine-tune client selection by considering data heterogeneity within selected clients and data plus system heterogeneity between selected and non-selected clients.  

\begin{figure}[t]
    \centering
    \includegraphics[width=1\linewidth]{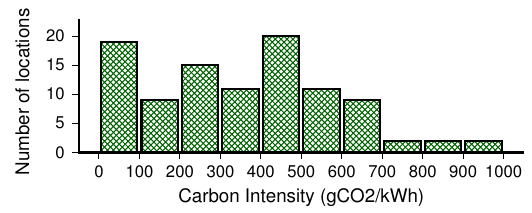}
    \vspace{-0.8cm}
    \caption{\textbf{\emph{Histogram of average carbon-intensity across 100 locations worldwide. Carbon-intensity ranges from near 0g$\cdot$CO$_2$g/kWh to nearly 1000g$\cdot$CO$_2$g/kWh.}}}
    \label{fig:carbon-distribution-locations}
   \vspace{-0.2cm}
\end{figure}

\noindent
\textbf{2 -- Spatiotemporal variations in carbon intensity.} 
FL's carbon footprint depends on energy's carbon intensity, measured in grams of \(\mathrm{CO_2}\) equivalent per kilowatt-hour (\ciunit), across geographically distributed clients. 
Due to differences in energy sources—ranging from fossil fuels like coal and gas to renewables like hydro and wind—energy's carbon intensity can vary significantly between client locations. 
Figure~\ref{fig:carbon-distribution-locations} shows the yearly average carbon intensity values from 100 different locations across the globe, obtained using ElectricityMap~\cite{electricitymaps}, a third-party service providing aggregated real-time carbon-intensity data of the grid's electricity supply. Intelligent carbon-aware client selection selects clients only if their carbon intensity is low, reducing the carbon cost incurred by each client and, ultimately, FL's global carbon footprint.

\noindent
\textbf{3 -- Resource-efficient FL for carbon optimizations.} 
As discussed, prior work defines resource efficiency in terms of the time to reach a target model accuracy. 
This may not always translate to reducing carbon footprint, as participating clients may have high carbon during that time. 
Also, the work on reducing wasted computations at stragglers, at best, reduces system-level carbon waste but does not impact the carbon emitted by each client. 
Prior work such as Oort can be extended to reduce carbon by treating carbon as a resource or a cost and maximizing 1/carbon instead of utility. 
However, in~\autoref{sec:evaluation}, we show that 
extending existing resource-efficient FL approaches for carbon awareness by simply changing their objective to carbon reduction does not result in a desirable tradeoff between carbon, training time, and model accuracy.

Optimizing FL's carbon footprint instead requires integrating carbon awareness not only in the decision of \emph{which} clients to select but also \emph{how many} to select and \emph{when}. We now present the FL model training insights that \systemName{} leverages to inform its decisions.

\begin{figure*}[t]
    \minipage{0.315\textwidth}
    \includegraphics[width=\linewidth]{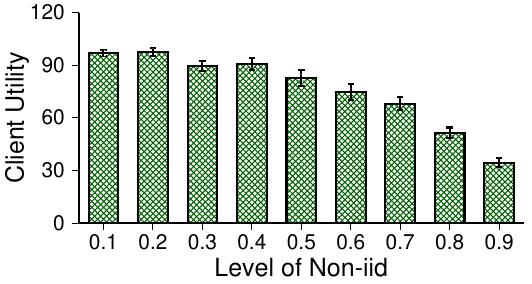}\vspace{-1em}
    \caption{Impact of data heterogeneity across clients on the statistical utility of a given client toward the global model.}\label{fig:non-iid-impact}
    \endminipage\hfill
    \minipage{0.315\textwidth}
    \includegraphics[width=\linewidth]{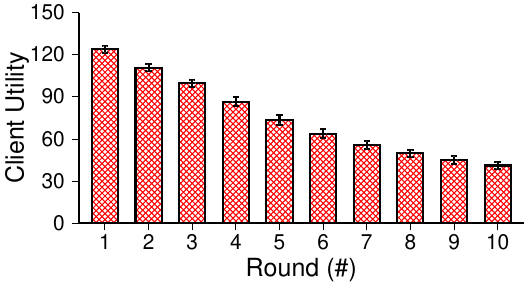}\vspace{-1em}
    \caption{Impact of the number of training rounds on the statistical utility of a client at a fixed non-iid level.}\label{fig:rounds-impact}
    \endminipage\hfill
    \minipage{0.315\textwidth}
    \includegraphics[width=\linewidth]{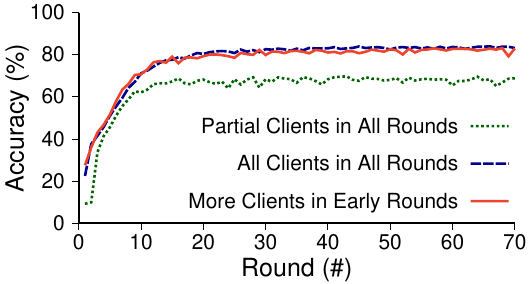}\vspace{-1em}
     \caption{Impact of how many clients are selected across rounds on the accuracy of the globally trained model.}\label{fig:clp-impact} 
    \endminipage
    \vspace{-0.3cm}
\end{figure*}

\subsection{Key Ingredients of \systemName}
\label{sec:motivation}
\systemName decomposes the problem of carbon-efficient client selection into three key steps: first, determining \emph{when} and \emph{how many} clients to select over time; second, deciding \emph{which} clients to select in each round; and third, provisioning \emph{how many} replica tasks to mitigate delays induced by stragglers. 

In making these decisions carbon-aware, \systemName exploits fundamental insights from prior work on how the utility of clients in improving global model accuracy can change based on data heterogeneity, the number of rounds progressed, and the number of clients selected in each round.

\vspace{0.1cm}
\noindent
\textbf{1 -- Variations in utility across clients.}
A client's statistical utility quantifies its importance in improving model accuracy~\cite{oort-osdi}.
\autoref{fig:non-iid-impact} shows a client's statistical utility as a function of data heterogeneity across clients (quantified using the level of \noniid).
In general, if all clients have the same distribution of classes, the data is independent and identically distributed (\iid). If clients have different distributions of classes, it is considered \noniid. 
\autoref{sec:evaluation} describes how we compute the level of \noniid for a given dataset. 

As shown in~\autoref{fig:non-iid-impact}, the average utility of participating clients decreases as the level of \noniid increases; model updates from clients with highly specific data are less useful.
Instead of blindly minimizing carbon cost, \systemName opportunistically trades carbon footprint for gain in model accuracy, i.e., it accepts a high carbon cost for a client with high utility. More formally, \emph{it selects clients based on their normalized utility per unit of carbon}. 

\noindent
\textbf{2 -- Change in utility over rounds.} 
If a model has learned from a given dataset, even partially, the value of learning on that data reduces. 
~\autoref{fig:rounds-impact} shows that as clients participate in training rounds, the global model learns from their data, and their statistical utility decreases. 
This is a useful property as it changes the order of clients; recently selected clients go down, and new clients with slightly smaller utility per unit of carbon have a higher ranking. 
This intrinsic shuffling of clients exposes \systemName to a broader range of clients, allowing it to select clients with lower carbon. 

\noindent
\textbf{3 -- Impact of time-varying number of clients.}
FL client selection approaches such as Oort and REFL select the same number of clients across all rounds. 
However, prior work on ML training in centralized and federated settings has highlighted the presence of an initial period when the model learns very fast, and exposure to a large number of high-quality clients can enable it to reach model accuracies that are not possible otherwise~\cite{clp_accordion,clp_fl}. 
This period is akin to, and gets its name from, the \emph{critical learning period}-(CLP) in humans, where children learn very fast. 
~\autoref{fig:clp-impact} shows that selecting more clients in the early rounds to exploit the CLP can result in the same accuracy as selecting all clients at all times with astronomical carbon costs. 
\systemName amplifies this effect by selecting high-utility clients during the initial rounds, offsetting the suboptimality of later selection based on utility per unit of carbon.

\systemName exploits low carbon periods to over-provision the number of clients in each round, accommodating stragglers and reducing training time at the lowest carbon possible in wasted computation.

\noindent
\textbf{Comparison with related FL frameworks.} 
\autoref{table:baseline-algs} shows that  \systemName distinguishes itself from other resource-efficient approaches in FL by integrating carbon awareness into multiple aspects of client selection, an aspect not handled by the prior work. In summary, \emph{\systemName selects high-utility but low-carbon clients during the early rounds and opportunistically launches replicas during low-carbon periods to mitigate stragglers.}
As shown in~\autoref{sec:evaluation}, \systemName significantly reduces the carbon footprint while maintaining the accuracy and training time of the carbon-agnostic policies.

\section{\systemName Design}
\label{sec:design}
\begin{figure}[t]
    \centering
    \includegraphics[width=1\linewidth]{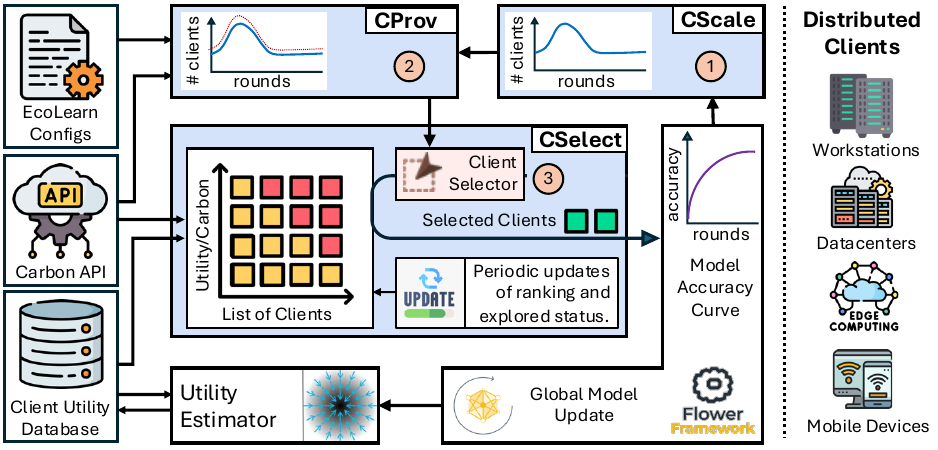}
  \vspace{-0.6cm}
    \caption{\systemName design and its components.}
    \vspace{-0.1cm}
    \label{fig:ecolearn-design}
\end{figure}

~\autoref{fig:ecolearn-design} provides an overview of \systemName's general design, which includes carbon-aware policies as a per-round client selection mechanism that can easily be extended to different federated learning (FL) frameworks. A design choice enabling integration with higher-level APIs that provide FL-based inference services. 

\systemName integrates with and extends existing FL frameworks that already provide basic functions of communication, client configurations, and model updates at the aggregator. 
However, FL frameworks do not provide native support for monitoring carbon footprint, or any other cost, of the clients. 
We had to implement the auxiliary modules needed for carbon awareness, such as the time- and location-specific carbon information services. 
As discussed in~\autoref{sec:eval_setup}, we chose the Flower Framework~\cite{flower} in our implementation, which is one of the most widely used FL frameworks for FL research and practical deployments. 
However, our design also applies to similar frameworks that enabling configuring the list of clients that are selected in each round of training. 

We next detail the \systemName modules and describe how they enable \systemName to make three decisions: 

\begin{enumerate}[leftmargin=*, itemsep=0.15cm, topsep=4pt]
    \item[\ding{182}] How many clients should it select in each round to exploit the learning benefit of early rounds? (\autoref{sec:cscale})

    \item[\ding{183}] Given the desired number of clients for a round, how many extra clients should it provision to ensure that the desired number of clients respond in time? (\autoref{sec:cprov})
    
    \item[\ding{184}] In selecting a specific number of clients, which specific clients should it select? (\autoref{sec:cselect})
\end{enumerate}

In detailing these modules, we highlight our design decisions to make these modules carbon-aware and the challenges we faced in practically implementing our solutions.

\subsection{\cscale: Time-varying Client Scaling}
\label{sec:cscale}

FL typically uses the same number of clients for all training rounds. 
However, prior work on analyzing learning patterns in ML and FL suggests that selecting more clients in the earlier training rounds is critical for creating strong connections that better understand the distribution of client data~\cite{Achille:2018:Critical, clp_accordion, clp_fl}.
As we discussed in~\autoref{sec:motivation}, and shown in~\autoref{fig:clp-impact}, the learning boost from a higher number of clients in earlier rounds enables the model to increase the accuracy achieved at the end of training~\cite{clp_accordion, clp_fl}. 
\systemName exploits this phenomenon to reduce the number of rounds and carbon emissions needed to reach a target accuracy. 

In practice, \emph{scaling the clients across rounds} to exploit the critical learning period is non-trivial, as when it starts, how long it lasts, and how many clients are enough during this period is not known \emph{a priori}. In addition, the answers to these questions change based on model- and data-specific factors. 
Therefore, the critical learning period must be identified online in a way that adapts to different FL settings and environments. 
To solve this problem, \systemName develops a mechanism to detect the start and end of the critical learning period and a set of policies that scale the number of clients.

\begin{algorithm}[t]
\small
\SetAlgoLined
\SetKwInOut{Input}{Input}
\SetKwInOut{Output}{Output}
\Input{ Scaling policy ($\mathcal{P}$), 
CLP detection threshold ($\tau_{\text{deriv}}$), 
minimum no. of clients ($\text{N}_{\text{min}}$)}
\Output{$\text{N}_{\text{CLP}}$}
A $\gets$ \texttt{AccuracyCurve()} \label{alg:line:1}
A'(r)  $\gets \frac{dA}{dr}$\; 
$\mu$ $\gets |A'(r)|$\;
$\mu_r \gets \frac{1}{v} \sum_{i=r-v+1}^{r} |A'(i)|$\; 
$\text{N}_{\text{CLP}} \gets \texttt{IncreaseClients}(\mathcal{P}) \ \textbf{if} \ \mu_r < \tau_{\text{deriv}} \ \textbf{else} \ \max(\texttt{DecreaseClients}(\mathcal{P}), N_{\text{min}})$ \\ \label{alg:line:4}
\Return{\textnormal{$\text{N}_{\text{CLP}}$}}
\caption{\texttt{ClientScaler}}
\label{alg:client-scaling-latex}
\end{algorithm}

\begin{figure}[t]
    \minipage{0.23\textwidth}
    \includegraphics[width=\linewidth]{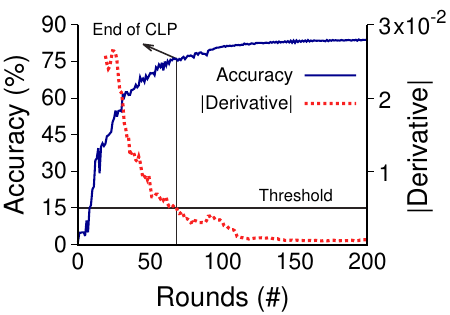}\vspace{-0.2cm}
    \caption{Finding the critical learning period (CLP).}\label{fig:finding-CLP-design}
    \endminipage
    \hfill
    \minipage{0.23\textwidth}
    \includegraphics[width=\linewidth]{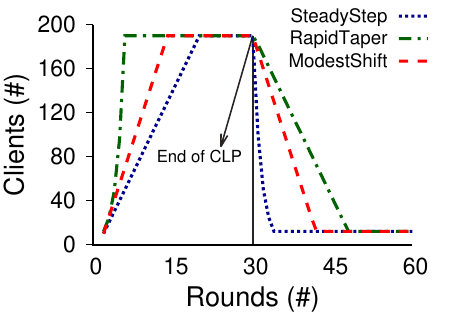}\vspace{-0.2cm}
     \caption{Variants of client scaling strategies.}\label{fig:CLP-client-strategies}
    \endminipage
    \vspace{-0.5cm}
\end{figure}

\vspace{0.1cm}
\noindent
\textbf{Detecting critical learning period.} 
Prior work analyzes the weighted aggregate of loss gradients across all the clients during their local training to identify when the learning slows down~\cite{clp_accordion, clp_fl}. 
However, this approach is intrusive and incurs the overhead of collecting gradients. 
To tackle this challenge, as stated in~\autoref{alg:client-scaling-latex}, \cscale uses the accuracy curve of the global model to mark the start and end of CLP (\autoref{alg:line:1}).
\autoref{fig:finding-CLP-design} shows the curves for both the accuracy (left $y$-axis) and the magnitude of its derivative (right $y$-axis). 
Typically, accuracy increases sharply during the initial learning phase, resulting in a high, rapidly increasing derivative, indicating the CLP. Gradually, the slope of the accuracy curve decreases, and its derivative starts approaching 0. As the magnitude of the derivative falls below a predefined threshold ($\tau_{\text{deriv}}$) set by \systemName, the detection algorithm marks that point as the conclusion of the CLP. 

\vspace{0.1cm}
\noindent
\textbf{Scaling the number of clients.} 
Once the CLP starts, the \texttt{ClientScaler} scales the number of clients until the magnitude of the derivative falls below the specified threshold; it starts decreasing the number of clients at that point (\autoref{alg:line:4}). 
\cscale enables three policies with different scale-up and down factors, as adjusting these factors affects the training performance and carbon cost.

    \vspace{0.05cm}
    \noindent
    \textbf{1 -- \texttt{SteadyStep}:} Clients are added steadily during CLP and dropped steeply after CLP. Each round adds $\alpha_1 \times N$ clients ($\alpha_1$ is a small value). Post CLP, $\beta_1 \times N$ clients are dropped ($\beta_1$ is relatively larger).

    \vspace{0.05cm}
    \noindent
    \textbf{2 -- \texttt{RapidTaper}:} Clients are added rapidly during CLP and dropped steadily after CLP. The values of $\alpha_2$ and $\beta_2$ are reversed, with $\alpha_2$ being large and $\beta_2$ being small.

    \vspace{0.05cm}
    \noindent
    \textbf{3 -- \texttt{ModestShift}:} Clients are added and dropped at a moderate rate, with $\alpha_3$ and $\beta_3$ being similar values.

Here, $\alpha_2 > \alpha_3 > \alpha_1$, conversely, $\beta_1 > \beta_3 > \beta_2$ and $N$ is the total number of clients.
~\autoref{fig:CLP-client-strategies} shows how the number of clients is scaled up and down based on these strategies. 
Note that \texttt{SteadyStep} minimizes the time during training when a higher number of clients are selected and thus reduces the carbon cost the most while maintaining the accuracy compared to the baseline policies. We thoroughly evaluate the effect of scaling factors on performance in~\autoref{sec:policy-variants}.

\setlength{\textfloatsepsave}{\textfloatsep} 
\setlength{\textfloatsep}{6pt} 
\begin{algorithm}[t]
\footnotesize
\SetAlgoLined
\SetKwInOut{Input}{Input}
\SetKwInOut{Output}{Output}

\Input{$\text{N}_\text{{CLP}}$, carbon weight ($\omega_{\text{carbon}}$), training time weight ($\omega_{\text{time}}$)}
\Output{CLP and staggler-aware (SA) number of clients $\text{N}_\text{{CLP--SA}}$}

over\_prov\_factor $\gets$ \texttt{CarbonTrainTimeTradeoff}($\omega_{\text{carbon}}$, $\omega_{\text{time}}$) \\ \label{alg2:line1}
$\text{N}_\text{{CLP--SA}}$ $\gets$ $\text{N}_\text{{CLP}}$ $\times$ (1 + over\_prov\_factor) \\
\Return{\textnormal{$\text{N}_\text{{CLP--SA}}$}}
\caption{\texttt{StragglerProvisioning}}
\label{alg:straggler-provisioning-latex}
\end{algorithm}

\vspace{-0.2cm}
\subsection{\cprov: Straggler-Aware Client Provisioning}
\label{sec:cprov}
The client scaling component discussed in the previous section aims to reduce the average number of clients per round, lowering carbon costs. 
However, selecting a smaller number of clients may lead to an unexpected drop in accuracy and an increase in training time if some of the selected clients are stragglers.
Recall from~\autoref{sec:motivation} that stragglers take longer to complete their training than non-stragglers, and thus, they can potentially slow down the overall FL training process. 
Thus, \systemName's second design component focuses on determining \emph{how many extra} clients we need to provision to mitigate the effect of stragglers. 

Existing straggler mitigation techniques use a static approach, over-provisioning a fixed number of clients per training round~\cite{oort-osdi, fedavg-mlsys, wiesner2024fedzero}. This over-provisioning creates a trade-off between carbon cost and training time: increasing replication reduces training time but raises carbon costs. However, beyond a certain threshold, adding clients yields diminishing returns in time reduction, severely limiting carbon savings. \systemName addresses this trade-off by adjusting the replication rate based on the carbon cost distribution across clients and over time (\autoref{sec:straggler-provisioning-evaluation}). When the carbon cost is high, \systemName selects fewer additional clients, and vice versa.

\autoref{alg:straggler-provisioning-latex} outlines this approach. \systemName configures the weights ($\omega_{\text{carbon}}$ and $\omega_{\text{time}}$) to balance the carbon cost of extra clients and the reduction in training time. A higher $\omega_{\text{carbon}}$ indicates a greater focus on minimizing carbon cost. 
Using these weights, \texttt{CarbonTrainTimeTradeoff()} (\autoref{alg2:line1}) determines the additional clients needed. The outcome is a carbon-aware selection of clients that optimizes the learning period while mitigating stragglers.

\vspace{-0.2cm}
\subsection{\cselect: Carbon-Aware Client Selection}
\label{sec:cselect}

The carbon-aware client selection component is a fundamental building block of \systemName, which aims to select clients with the lowest carbon cost and highest utility to achieve high model accuracy. 
In designing \systemName's client selection policy, the carbon cost is based on computing's carbon intensity (\cciunit):
\[
\text{CI}_{\text{comp}} = \text{CI}_{\text{energy}} / \text{EE}_{\text{comp}}
\]
where \(\text{CI}_{\text{energy}}\) is the carbon intensity of energy (\ciunit) and \(\text{EE}_{\text{comp}}\) represents a client's computational energy efficiency (\efunit).  Thus, computing's carbon intensity depends on both energy's carbon intensity and each client's energy-efficiency. \systemName estimates the current carbon cost when determining which clients to select.

\begin{algorithm}[t]
\small
\SetAlgoLined
\SetKwInOut{Input}{Input}
\SetKwInOut{Output}{Output}
\Input{CLP and straggler-aware number of client (\textnormal{$\text{N}_\text{{CLP--SA}}$}), exploration factor ($\epsilon$), carbon weight ($\omega_{\text{carbon}}$), utility weight ($\omega_{\text{util}}$) } 
\Output{Final set of client $\mathbf{S}$}

$\lambda$ $\gets$ \texttt{UtilityCarbonRanking}($\omega_{\text{carbon}}$, $\omega_{\text{util}}$)  \label{alg3:line1}

$\mathbf{S}_\text{{explore}}$ $\gets$ \texttt{GetUnexploredClients}($\epsilon $ $\cdot$ \textnormal{$\text{N}_\text{{CLP--SA}}$}, $\lambda$) \label{alg3:line2} \\

$\mathbf{S}_\text{{exploit}}$ $\gets$ \texttt{GetExploitationClients}(($1 -\epsilon$) $\cdot$ \textnormal{$\text{N}_\text{{CLP--SA}}$}, , $\lambda$) \label{alg3:line3} \\

$\mathbf{S}$ $\gets$ $\mathbf{S}_\text{{explore}}$ + $\mathbf{S}_\text{{exploit}}$ \\
\Return{\textnormal{$\mathbf{S}$}}
\caption{\texttt{ClientSelection}}
\label{alg:client-selection-latex}
\end{algorithm}

We devise two baseline carbon-aware client selection policies:

\noindent
\textbf{1 -- Carbon-based (\texttt{carbon}).}
A naive client selection policy selects \( \mathcal{K} \) clients in each round with the lowest carbon cost. While this policy provides the maximum carbon savings, it ignores clients' utility in improving the model performance during client selection, significantly increasing training time and lowering accuracy due to insufficient data coverage. 

\noindent
\textbf{2 -- Carbon- and utility-based (\texttt{carbon+utility}).} 
Although the carbon-based approach effectively reduces the overall carbon footprint of training, selecting clients solely based on carbon can introduce bias~\cite{fairfl}, impacting the overall model accuracy. Instead, we select clients in each round based on their normalized utility~\cite{oort-osdi} per unit carbon. We call this policy a ``\texttt{utility+carbon}'' aware client selection policy. 

For a given client \( k \), we define this efficiency metric as: 
\[
E_k =\frac{U_k}{C_k} = \frac{|D_k|}{C'_k} \times \sqrt{\frac{1}{|D_k|} \sum_{i \in D_k} \text{Loss}(i)^2}.
\]
where, \( U_k \) is the statistical utility, and \( C'_k \) represents the instantaneous computing's carbon intensity associated with client \( k \) at location \( x \). \( |D_k| \) is the total number of training samples from client \( k \), and \( \text{Loss}(i) \) is the training loss for sample \( i \). Clients with higher accumulated losses are considered more significant for future rounds.

This policy outperforms the accuracy of a simple \texttt{carbon}-based approach by incorporating utility alongside carbon cost when selecting clients. While it effectively reduces the overall carbon footprint of FL training, the potential exclusion of high-utility, high-carbon clients limits model accuracy compared to carbon-agnostic utility-based approaches, which may represent an acceptable but undesirable trade-off. Thus, simply making existing carbon-agnostic utility-based techniques carbon-aware is insufficient, as it significantly reduces accuracy (shown in~\autoref{sec:cost-aware}). This limitation motivates \systemName’s additional optimizations, which are discussed below.

\begin{algorithm}[t]
\small
\caption{\systemName Framework}
\label{alg:ecolearn-framework}
\KwIn{clp policy $\mathcal{P}$, exploration factor $\epsilon$, stopping criteria $\Lambda$, CLP threshold $\tau_{\text{deriv}}$, carbon cost weight $\omega_{\text{carbon}}$, training time weight $\omega_{\text{time}}$, utility weight $\omega_{\text{util}}$, minimum clients $N_{\text{min}}$}
\KwOut{final model parameters $\Theta_{\text{final}}$}

\While{$\Lambda \neq \text{true}$}{ 
    $\text{N}_\text{{CLP}}$ $\gets$ \texttt{ClientScaling}($\mathcal{P}$, $\tau_{\text{deriv}}$, $N_{\text{min}}$) \\
    $\text{N}_\text{{CLP--SA}}$ $\gets$ \texttt{StragglerProv}($\text{N}_\text{{CLP}}$, $\omega_{\text{carbon}}$, $\omega_{\text{time}}$) \\
    $\mathbf{S}$ $\gets$ \texttt{ClientSelection}($\text{N}_\text{{CLP--SA}}$, $\omega_{\text{carbon}}$, $\omega_{\text{util}}$, $\epsilon$) \\
    \texttt{StartTraining}(clients\_round) \\
    $n \gets 0$ \\
    \While{$n < \text{N}_\text{{CLP}}$}{
        $n \gets$ \texttt{ClientsFinishTraining}()
    }
    $\Theta_{\text{final}} \gets$ \texttt{UpdateModelParams}($n$) \\
}
\Return{$\Theta_{\text{final}}$}
\end{algorithm}

\subsection{\systemName's Policy} 
\systemName's policy combines the functionality of all the previous modules, as described in~\autoref{alg:ecolearn-framework}.

At the start of training, \systemName's client selection policy initiates by selecting clients in the first round to calculate utility values, which are then stored in a client utility database. 
The process begins by selecting \(\text{M} \times \text{N}\) clients, where \(\text{M}\) is a fraction in the range \([0,1]\), and \(\text{N}\) denotes the total number of clients. 
In subsequent rounds, \systemName receives a list of clients (\(\lambda\)) and ranks them based on \( E_k \)~\autoref{alg:ecolearn-framework}-\autoref{alg3:line1}. Since the utility \(U_k\) is determined only after a client participates in training, sampling all clients individually becomes impractical for large numbers. To address this, \systemName uses an \emph{exploration-exploitation technique}, balancing between utilizing clients with known utilities (exploitation) and exploring new clients to identify those with high utility (exploration). The system selects (exploits) \((1 - \epsilon) \times \text{N}_\text{CLP--SA}\) clients from the already explored clients in the \(\lambda\) list~(\autoref{alg3:line3}), while the remaining \(\epsilon \times \text{N}_\text{CLP--SA}\) clients are selected from the unexplored client set~(\autoref{alg3:line2}). Here, \(\epsilon\) takes values in the range \([0,1]\), controlling the rate of exploration.

Simultaneously, \cscale is monitoring for the start of CLP. If the CLP has not yet started, it returns a default value for the number of clients (i.e., \(\text{M} \times \text{N}\)). Once CLP starts, it uses the scaling policies to determine the number of clients for the next round. 
In either scenario, the number of clients provided by the \cscale module is passed to the \cprov module, which adds the extra number of clients based on the value it received (N$_{\text{CLP}}$) and the average carbon intensity across clients. 
The final number of clients is then passed to the \cselect module, which selects specific clients based on their normalized utility per carbon values while accounting for the need to explore new clients for utility estimation. 
As discussed in~\autoref{sec:motivation}, the utility of the clients that have already participated in the training process decreases over time, and the unexplored clients are highly likely to have higher utility, which helps with selecting high-value clients at the start. This also ensures that \systemName does not repeatedly select the same clients during the training process, which could introduce bias and negatively impact model performance.
For each round, we get the final set of clients that should participate in the training process. These rounds continue until the global model reaches the maximum training time threshold or hits the target accuracy on the test dataset.

\section{\systemName Evaluation}
\label{sec:evaluation}
This section presents our evaluation methodology (\autoref{sec:eval_setup}), evaluates \systemName's three core design components (\autoref{sec:cost-aware} to \autoref{sec:policy-variants}), and analyzes \systemName's generalizability across real-world datasets (\autoref{sec:generalization}).

\subsection{Evaluation Methodology}
\label{sec:eval_setup}

We present our evaluation methodology, covering (1) implementation and compute resources, (2) training datasets and models, (3) carbon cost traits, and (4) baseline policies.

\noindent\textbf{Implementation.} 
Our implementation extends the Flower framework~\cite{flower}, which emulates the communication between a global server and geographically distributed clients. Flower uses PyTorch as its underlying ML library. We leverage Flower's Virtual Client Engine (VCE) to configure and manage resources, enabling efficient FL execution. The virtual clients are implemented using Ray~\cite{ray}. 
We implement auxiliary modules to support carbon- and utility-aware client selection, carbon-aware straggler provisioning, and time-varying client scaling, based on \systemName’s design (\autoref{sec:design}).  Our implementation adds approximately 2000 lines of code, and we plan to release the code publicly.

\begin{table}[t]
    \centering
    \footnotesize
    \caption{\textbf{\emph{Statistics of datasets used in the evaluation.}}}
    \vspace{-0.3cm}
    \label{tab:dataset-stats}
    \begin{tabular}{|c|c|c|c|}
    \hline
    \textbf{Datasets} & \textbf{Samples} &\textbf{Emulated }  & \textbf{Inherently}   \\ 
    \textbf{Datasets} & \textbf{} &\textbf{Clients ($N$)}  & \textbf{\noniid}   \\ \hline \hline
    EMNIST           & 814,255          & 1,000     & No              \\ \hline
    Shakespeare      & 422,615          & 660       & Yes             \\ \hline
    Google Speech    & 105,829          & 100       & Yes             \\ \hline
    \end{tabular}%
\end{table}

\vspace{0.05cm}
\noindent\textbf{Experimental setup.} We evaluate \systemName on a cluster of 40 \texttt{NVIDIA GTX 1080Ti} and 6 \texttt{NVIDIA RTX 8000} GPUs to emulate a medium-to-large-scale FL scenario. We emulate up to 1,000 participants in the FL training, representing large-scale edge deployments using Flower~\cite{flower}. Our emulation and evaluation setup is comparable to or larger than ones used in prior work on optimizing FL training~\cite{oort-osdi,wiesner2024fedzero,refl-eurosys,clp_fl}.

\vspace{0.05cm}
\noindent\textbf{Datasets and models.}
We use three datasets in our experiments, each with distinct scale and complexity: EMNIST~\cite{emnist-dataset}, Shakespeare~\cite{caldas2018leaf}, and Google Speech Commands~\cite{warden2018speech}. \autoref{tab:dataset-stats} summarizes their characteristics. EMNIST, a vision dataset for image recognition, Shakespeare contains dialogues for next-character prediction, and Google Speech Commands for speech recognition tasks. 

\vspace{0.05cm}
\noindent
\textbf{Inducing stragglers.}
Recalling \autoref{sec:motivation}, stragglers impact FL's efficiency by increasing training time. To enable realistic FL evaluation and simulate real-world delays, we induce stragglers in our system using heavy-tailed distributions~\cite{behrouzi2020efficient}, meaning some clients fall in the heavy tail and experience more delay than others.

\vspace{0.05cm}
\noindent
\textbf{Achieving \noniid data distributions. }
We leverage the Fang distribution~\cite{fang2020local} to precisely control and quantify the impact of data heterogeneity (\noniid\emph{ness}). The level of ``\noniid\emph{ness}'' is controlled using a parameter, $\kappa$, which ranges from 0 (fully i.i.d) to 1 (fully \noniid). At $\kappa = 1$, a single client contains only one class with no representation of other classes. At $\kappa = 0$, all clients hold local data of all classes in equal proportions. 
For $0 < \kappa < 1$, the degree of \noniid\emph{ness} ranges from mild to extreme \noniid. 

\begin{figure}[t]
    \centering
    \begin{tabular}{@{}c@{\hspace{-0.5mm}}c@{}}
    \includegraphics[width=0.51\linewidth]{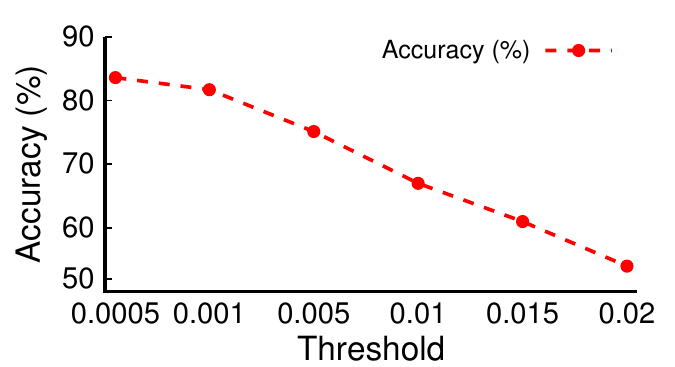} & 
    \includegraphics[width=0.51\linewidth]{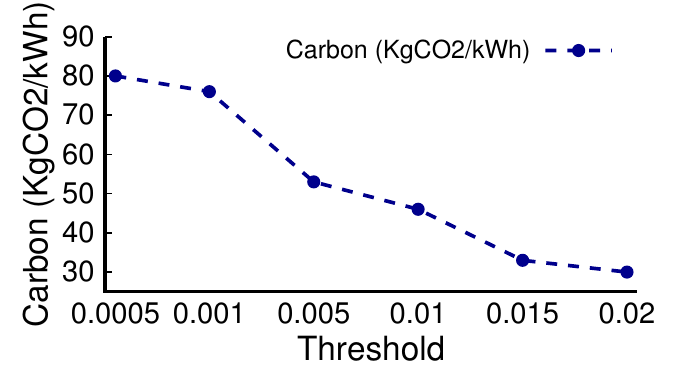}  \\
    (a) Accuracy & (b) Carbon Cost
    \end{tabular}
    \vspace{-0.3cm}
    \caption{\textbf{\emph{Effect of varying detection threshold.}}}   
    \label{fig:detection-threshold}
\end{figure}
 
\begin{figure*}[t]
    \minipage{0.322\textwidth}
    \vspace{1.1em}
    \includegraphics[width=\linewidth]{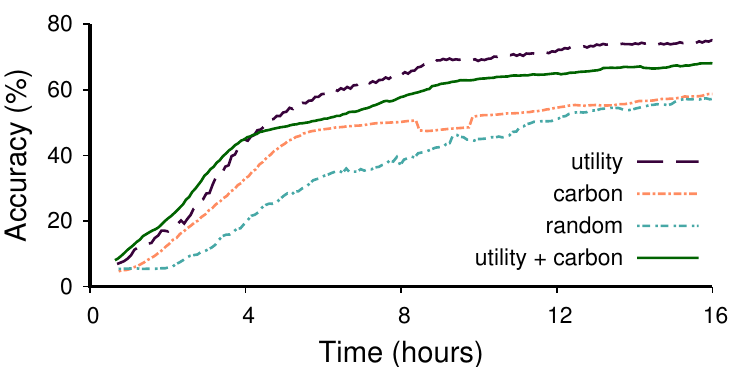}\vspace{-1em}
    \caption{Accuracy curves for baseline and \systemName's policies for the EMNIST dataset (Higher accuracy is better).}\label{fig:accuracy-emnist-all-policy}
    \endminipage\hfill
    \minipage{0.322\textwidth}
    \includegraphics[width=\linewidth]{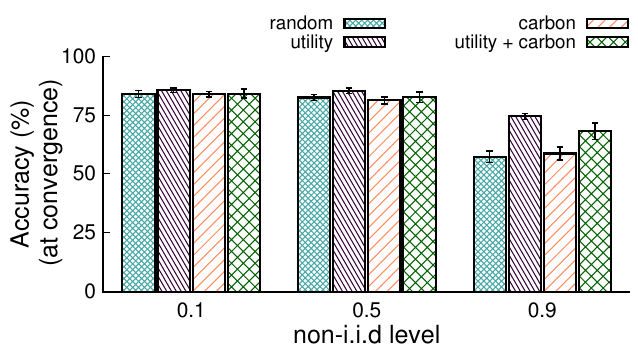}\vspace{-1em}
    \caption{Effect of \noniid levels on the performance of the global model under different client selection policies.}\label{fig:accuracy-emnist-policy}
    \endminipage\hfill
    \minipage{0.322\textwidth}
    \includegraphics[width=\linewidth]{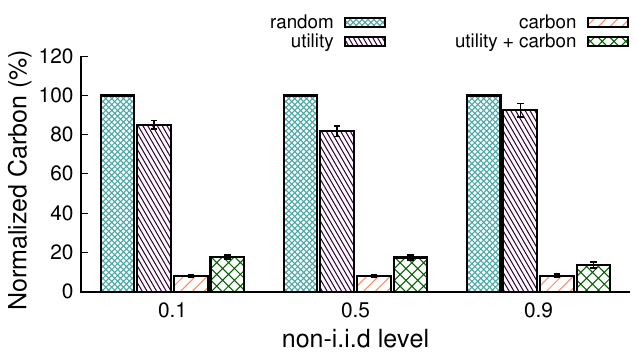}\vspace{-1em}
     \caption{Carbon costs for client selection policies, plotted in comparison with the random baseline. (Lower \% is better)}\label{fig:convergence-rounds-carbon} 
    \endminipage
    \vspace{-0.3cm}
\end{figure*}

\vspace{0.05cm}
\noindent
\textbf{Detection threshold ($\tau_{\text{deriv}}$) in CLP. }
As mentioned in \autoref{sec:cscale}, if the magnitude of the derivative falls below a threshold ($\tau_{\text{deriv}}$), \systemName{} marks that point as the conclusion of CLP. Changing the value of this threshold impacts both the accuracy and the carbon cost of training. As $\tau_{\text{deriv}}$ increases, CLP consists of fewer rounds, causing it to shrink. Consequently, both accuracy and carbon footprint decrease, as \systemName{} utilizes a larger number of clients in fewer rounds. Conversely, a smaller $\tau_{\text{deriv}}$ results in better accuracy at the cost of higher carbon emissions. In our experiments, we fine-tune $\tau_{\text{deriv}}$ within the range $0.0005 < \tau_{\text{deriv}} < 0.03$. Based on the observations in \autoref{fig:detection-threshold}, we set $\tau_{\text{deriv}}$ to 0.005.

\vspace{0.05cm}
\noindent
\textbf{Client scaling factors. }
In \autoref{sec:cscale}, we discussed three client scaling policies that vary the number of clients during and after CLP. We follow the pattern $\alpha_2 > \alpha_3 > \alpha_1$ and $\beta_1 > \beta_3 > \beta_2$, setting the scaling factor values as: $\alpha_1 = 0.01$, $\alpha_2 = 2$, $\alpha_3 = 0.015$, and $\beta_1 = 2$, $\beta_2 = 0.01$, $\beta_3 = 0.015$. These scaling factors (inspired by prior work~\cite{clp_accordion, clp_fl}) simulate different scaling behaviors, including steady scale-up, rapid scale-down, rapid scale-up, steady scale-down, and moderate scale-up and scale-down respectively.

\vspace{0.05cm}
\noindent
\textbf{Model parameters.}
We train ResNet-18~\cite{he2016deep} on EMNIST for image classification, a two-layer LSTM~\cite{li2020federated} on Shakespeare for next character prediction, and KWT-1~\cite{berg2021keyword} on Google Speech for speech classification. Initial learning rates were 0.01 for EMNIST, 0.001~\cite{berg2021keyword} for Google Speech, and 0.8~\cite{li2020federated} for Shakespeare, with batch sizes of 20 for EMNIST and Shakespeare, and 16 for Google Speech.

We use FedAvg~\cite{federated-learning-paper} as the aggregation strategy across all datasets and selection policies as it is the commonly used in practice and in baseline models. Other techniques like SCAFFOLD~\cite{karimireddy2020scaffold} or FedProx~\cite{li2020federated} can be used in future work. 

\smallskip
\noindent\textbf{Baseline policies.}
Below, we present the baseline approaches that we compare \systemName with.

\begin{enumerate}[leftmargin=*, topsep=0.1cm, itemsep=0.1cm]
    \item \textbf{Random selection.} Random client selection~\cite{fedavg-mlsys, MLSYS2022_a8bc4cb1} ensures fairness and reduces bias. However, it extends training time by including clients with less value or noisy data, leading to slower learning and more rounds to reach the desired accuracy.

    \item \textbf{Guided selection - Oort.} As mentioned in \autoref{sec:resource-efficient-fl} Oort~\cite{oort-osdi} employs a guided participant selection approach to prioritize clients with higher statistical utility to maximize system efficiency. This approach enhances time-to-accuracy by utilizing client data, significantly improving model performance and outperforming random selection. 
\end{enumerate}

To mitigate stragglers' induced delays, the baselines over-provision clients using a 1.3\( \mathcal{K} \) rule. Thus, provisioning 30\% more clients than the \( \mathcal{K} \) required, training all 1.3\( \mathcal{K} \) clients, and collecting updates from the first \( \mathcal{K} \) clients that respond the fastest.

\smallskip
\noindent\textbf{Evaluation Metric.}
To evaluate the performance and carbon footprint of carbon-aware and carbon-agnostic selection policies, we train them for the same duration and compare their accuracy and carbon cost. This approach offers a better evaluation metric than the time to reach a target accuracy, as it avoids infinite carbon costs and training times for policies that never reach the target. We use the utility-based baseline as a benchmark, training all policies for a time \( T_u \), the convergence time for the utility-based approach. We report results across 10 runs of each experiment. 

\subsection{Carbon-Aware Client Selection}
\label{sec:cost-aware}
\noindent
Here, we evaluate the performance and carbon cost of carbon-aware (\texttt{carbon-} and \texttt{utility+carbon based}) policies against carbon-agnostic (random and utility-based) baselines. These policies exclude the CLP and carbon-aware straggler mitigation, which are discussed in our final policy in \autoref{sec:policy-variants}.

\noindent
\textbf{Accuracy.} \autoref{fig:accuracy-emnist-all-policy} shows the accuracy for the above-discussed selection policies at a \noniid level of 0.9, with the x-axis representing wall-clock time (in hours). All policies are trained for the same duration—16 hours, equivalent to the utility-based selection's convergence time. Baseline strategies—utility-based and random selection—are shown by dashed magenta and dotted blue lines. Carbon-aware policies, \texttt{carbon}-based and \texttt{utility\-+carbon}, are represented by dotted orange and solid green lines, respectively.

At 0.9 \noniid, \autoref{fig:accuracy-emnist-policy} shows \texttt{utility+carbon} outperforms random and \texttt{carbon}-based approaches by selecting high-value, low carbon cost clients. Utility-based selection achieves the highest accuracy, surpassing \texttt{utility+carbon} by \emph{6.4\%} and random and \texttt{carbon}-based approaches by approximately \emph{17\%}. \texttt{utility+carbon} achieves approximately \textbf{\emph{10\%}} higher accuracy than \texttt{carbon}-based and random baselines. This result shows that by considering both utility and carbon costs in selection, the model performs better than just randomly selecting clients. However, it is still not as effective as purely utility-based selection because the trade-off between utility and carbon cost leads \systemName to select clients that may not maximize model performance (accuracy) but do reduce overall carbon emissions. We will see in \autoref{sec:policy-variants} that varying the number of clients in the CLP can close this accuracy gap between \texttt{utility+carbon} and utility-based selection and achieve, approximately the same accuracy as utility-based selection.

\noindent
\textbf{Carbon cost.}
\autoref{fig:convergence-rounds-carbon}, with a \noniid level of 0.9, shows the carbon cost for training time equivalent to the utility-based selection's convergence time. The y-axis shows the carbon costs of client selection policies in comparison to random selection. 
The \texttt{utility+carbon} outperforms carbon-agnostic policies, reducing carbon consumption by \textbf{\emph{86.4\%}} compared to random and by \textbf{\emph{85.3\%}} compared to utility-based selection. While \texttt{carbon} based approach emits the least carbon, it suffers a significant performance loss.

\noindent
\textbf{Effect of data heterogeneity.}
\autoref{fig:accuracy-emnist-policy} shows the effect of varying \noniid levels on client selection policies. At low \noniid values (0.1), the final model accuracy is similar across all selection policies, as the data is uniformly distributed among all clients, leading to similar contributions towards the model's performance. As the \noniid level increases to 0.5, differences in accuracy become more pronounced. The utility-based approach performs best in terms of accuracy by prioritizing clients with higher statistical utility, due to the pronounced effect of \noniid data across clients and some clients contributing more towards the performance of the model than others. However, in terms of carbon savings, the carbon-aware selection policies perform better as they prioritize selecting clients based on carbon, giving more weight to the carbon savings than to the accuracy. The \texttt{carbon} based approach reduces carbon cost the most (\emph{92\% less than the random baseline}). The \texttt{utility+carbon} approach achieves substantial savings of $\approx$\emph{79\%} and \emph{82.6\%} compared to utility-based and random-based, respectively.

\begin{figure*}[t]
    \minipage{0.322\textwidth}
    \includegraphics[width=\linewidth]{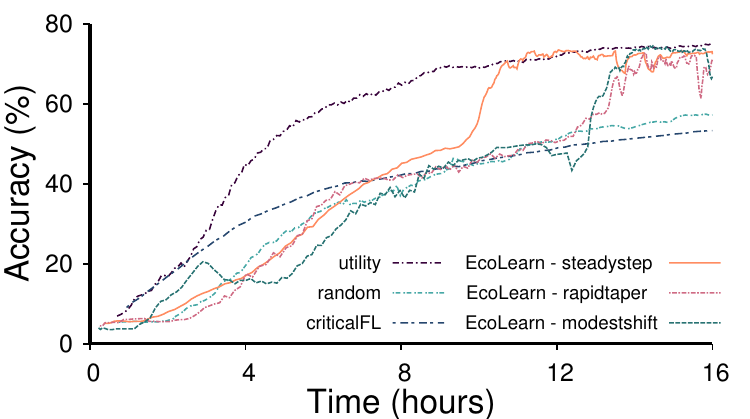}\vspace{-1em}
    \caption{\systemName's scaling strategies vs. baselines}\label{fig:CLP-final-model-accuracy-strategies}
    \endminipage\hfill
    \minipage{0.322\textwidth}
    \includegraphics[width=\linewidth]{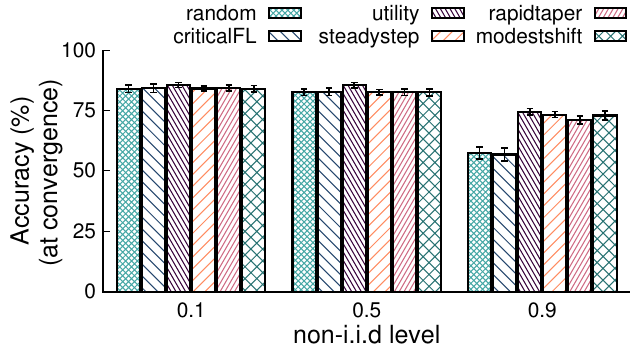}\vspace{-1em}
    \caption{Impact of \noniid on the accuracy of \systemName's scaling strategies}\label{fig:CLP-model-convergence-accuracy}
    \endminipage\hfill
    \minipage{0.322\textwidth}
    \includegraphics[width=\linewidth]{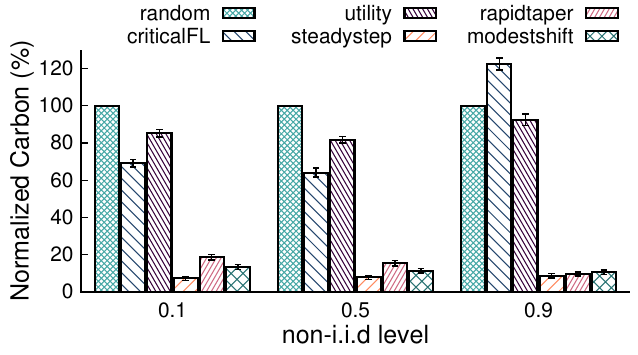}\vspace{-1em}
     \caption{Carbon cost of \systemName's scaling strategies compared to baselines}\label{fig:CLP-cost-to-random-convergence} 
    \endminipage
    \vspace{-0.4cm}
\end{figure*}

At \noniid level of 0.9, the effect of \noniid data across clients in-terms of clients' utility is more prominent. We observe that \texttt{utility+carbon} performs \textbf{\emph{10\%}} better than random and \texttt{carbon} based selections in accuracy, while giving \emph{6.4\%} less accuracy than utility-based selection. Additionally, \autoref{fig:convergence-rounds-carbon} shows that \texttt{utility+carbon} reduces carbon consumption by \textbf{\emph{86.4\%}} compared to random and \textbf{\emph{85.3\%}} compared to utility-based selection. 

\noindent
\textbf{Key point.} \emph{Our results show that \texttt{utility+carbon} policy of \systemName consistently outperforms random and utility-based selection baselines in terms of carbon cost across all the levels of \noniid while maintaining the accuracy of the model.}

\subsection{Straggler-Aware Client Provisioning}
\label{sec:straggler-provisioning-evaluation}
\begin{figure}[t]
    \centering
    \includegraphics[width=0.95\linewidth]{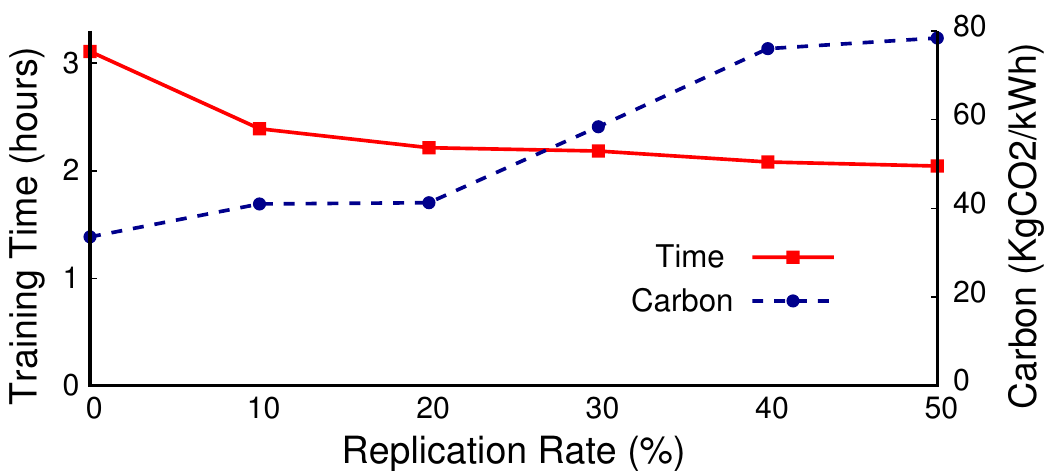}
    \vspace{-0.3cm}
        \caption{\textbf{\emph{Impact of client replication on training time and carbon cost for straggler mitigation.}}}
    \label{fig:straggler-time-and-carbon}
   \vspace{-0.4cm}
\end{figure}

Recalling \autoref{sec:cprov}, we over-provision clients in each training round to mitigate the effect of stragglers on training time. Our design navigates the trade-off between wasted carbon cost due to stragglers and reduced training time due to straggler mitigation. To analyze this trade-off, we compare the behavior of training time and carbon cost for FL at different levels of replication to mitigate delays induced by stragglers as shown in~\autoref{fig:straggler-time-and-carbon}. The x-axis represents the replication rate, the left y-axis shows the total training time, and the right y-axis illustrates the carbon cost of training across various replication rates. At 0\% replication, the training time is large due to the presence of stragglers, but carbon cost remains low, since no extra clients are over-provisioned. Higher replication rates (50\%) reduce training time by \emph{33\%} but increase carbon cost by over \emph{100\%}. Beyond a certain replication rate, training time decreases marginally due to the heavy-tailed distribution of stragglers.

\begin{figure}[t]
    \centering
    \includegraphics[width=0.95\linewidth]{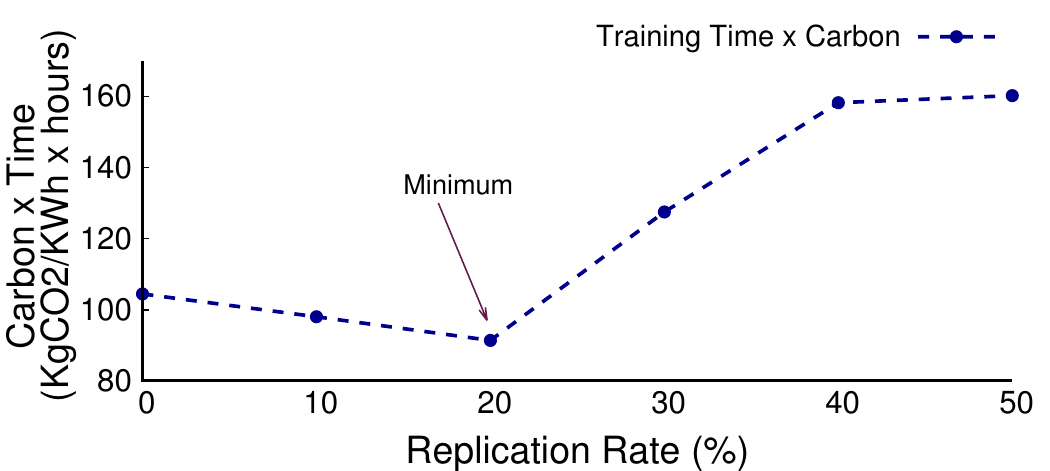}
    \vspace{-0.3cm}
    \caption{\textbf{\emph{Carbon$\times$time minima shows the optimal trade-off between carbon and training time across replication rates. }}}
    \label{fig:straggler-optimizing}
    \vspace{-0.2cm}
\end{figure}

Based on these observations, we optimize the carbon-time metric with respect to replication rate, indicating there is a specific replication rate that maximizes the benefit of reducing carbon cost relative to the overall training time. \autoref{fig:straggler-optimizing} shows the relationship between carbon-time and the replication rate, highlighting a minimum at \textbf{\emph{"20\% replication rate"}} (equivalent to 1.2\( \mathcal{K} \)). This replication rate provides the most effective balance between reducing carbon cost and improving the training time. \emph{Our evaluation demonstrates that the optimal replication rate is strongly influenced by the clients' carbon cost distribution.} Thus optimizing the replication rate based on the carbon cost distribution can yield significant improvements in reducing the carbon cost and the training time, than to use a static 1.3\( \mathcal{K} \) straggler over-provisioning factor. 

\subsection{Time-Varying Client Scaling}
\label{sec:policy-variants}
Recalling \autoref{sec:cost-aware}, \texttt{utility+carbon} outperforms random selection but is less effective than utility-based selection due to its trade-off between utility and carbon cost, leading to client selection that reduces carbon cost but does not maximize accuracy. In this section, we present that by varying the number of clients within and after CLP, we can increase the accuracy and further reduce the carbon cost. Additionally, we introduce another baseline approach, criticalFL \cite{clp_fl}, which dynamically detects CLP in the FL training process and adaptively determines the number of clients to participate in each FL training round. In contrast to \systemName's selection policy, which selects clients based on their utility and carbon cost in each round, criticalFL chooses clients randomly, without considering their utility or carbon cost.

\noindent
\textbf{Evaluating \systemName's client selection policy.}
Based on the insights about the carbon-aware straggler mitigation technique discussed in \autoref{sec:straggler-provisioning-evaluation}, we adopt a 1.2K replication rate (provisioning 20\% more clients and collecting results from the first K clients) for \systemName's CLP-based policies. This configuration avoids stragglers during FL training and optimizes the associated training time and carbon cost. In \autoref{fig:CLP-final-model-accuracy-strategies}, we compare \systemName's CLP-based client selection policies to the carbon-agnostic baseline policies. The y-axis represents the performance of all the selection policies at a \noniid level of 0.9, when trained for the same duration—16 hours, equivalent to the utility-based selection's convergence time. This avoids infinite carbon emissions and training times for policies that never reach benchmark accuracy. We observe that the accuracy of \systemName's CLP-based client selection policy (\emph{SteadyStep}) reaches \textbf{\emph{73.3\%}}, which is an improvement of \emph{3.5\%} compared to the non-CLP version of this policy (\texttt{utility+carbon} technique, discussed in \autoref{sec:cost-aware}). \systemName's selection policy is also comparable to utility-based selection (\emph{74.5\%}) and performs \textbf{\emph{11.2\%}} and \textbf{\emph{16.5\%}} better than random and criticalFL, respectively. These results are shown in \autoref{fig:CLP-model-convergence-accuracy} at a \noniid level of 0.9.

\autoref{fig:CLP-final-model-accuracy-strategies} illustrates an increase in accuracy when the critical learning period ends after around 9 hours of training, with the slope of this increase depending on the decaying factor by which we reduce the number of clients once the CLP ends. This behavior likely arises due to a very high \noniid level and the abrupt change in the number of clients from a large to a small number after CLP. Intuitively, learning becomes more challenging when many clients are provisioned within CLP and becomes easier once CLP ends, as there are fewer clients with fewer class representations, thus boosting accuracy. This behavior has been reported in~\cite{yan2021critical} and~\cite{smith2017don}.

\autoref{fig:CLP-cost-to-random-convergence} shows that, at a \noniid level of 0.9, \systemName's \emph{SteadyStep} policy reduces carbon cost by \textbf{\emph{91.5\%}}, \textbf{\emph{90.8\%}}, and \textbf{\emph{93\%}} compared to the random baseline, utility-based selection, and criticalFL, respectively, while maintaining comparable accuracy. Additionally, \systemName's \emph{SteadyStep} selection policy, integrated with CLP and carbon-aware straggler mitigation, reduces carbon cost by \textbf{\emph{37.5\%}} and increases accuracy by \textbf{\emph{5.2\%}} compared to its non-CLP version (\texttt{utility+carbon}) when trained for the same duration.

\vspace{0.1cm}
\noindent
\textbf{Effect of scaling strategies in CLP.}
\autoref{fig:CLP-model-convergence-accuracy} shows, \systemName's CLP based policies perform comparably in terms of accuracy across all levels of \noniid. However, \autoref{fig:CLP-cost-to-random-convergence} shows that the \systemName's CLP based policies significantly reduce the carbon cost of training across all \noniid levels compared to the baselines. 

Among \systemName's CLP based policies, \emph{SteadyStep} achieves the greatest reduction in carbon cost by minimizing the duration for which a large client base is active during the training period. As illustrated in \autoref{fig:CLP-client-strategies}, \emph{SteadyStep} strategy gradually increases the number of client during CLP and quickly drops them afterward, significantly reducing the carbon costs. In contrast, \emph{RapidTaper} incurs higher carbon costs due to its rapid increase in the number of clients during CLP and maintaining a large client base for an extended period. The \emph{ModestShift} strategy takes a moderate approach in client ramp-up during CLP and reduction post-CLP, resulting in higher carbon costs compared to \emph{SteadyStep}, but lower or comparable costs to \emph{RapidTaper} across all \noniid levels.

\begin{figure}[t]
    \centering
    \includegraphics[width=0.95\linewidth]{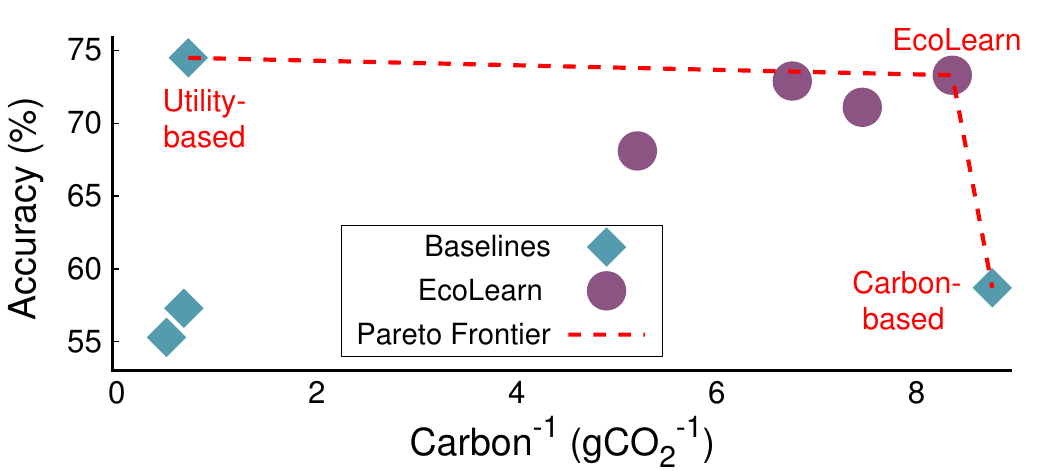}
    \vspace{-0.3cm}
    \caption{\textbf{\emph{Pareto-frontier of carbon cost vs. accuracy. Circles show \systemName's policies, stars indicate baselines. }}}
    \label{fig:pareto-front-accuracy-carbon}
 \vspace{-0.1cm}
\end{figure}

\systemName's \emph{SteadyStep} policy outperforms not only \emph{RapidTaper} and \emph{ModestShift} but all the other non-CLP selection policies and baselines. When trained for the same duration, the carbon cost for \emph{SteadyStep} is $\textbf{10.8}$$\times$, $\textbf{11.7}$$\times$, and $\textbf{14.4}$$\times$ less compared to random, utility-based selection, and criticalFL, respectively.

\vspace{0.1cm}
\noindent \textbf{Pareto-frontier.} 
\autoref{fig:pareto-front-accuracy-carbon} illustrates the Pareto frontier for the trade-off between accuracy (y-axis) and the inverse of carbon cost (x-axis). We aim for maximizing both our objective functions as we want to determine which policies lie on the Pareto frontier by maximizing the accuracy and the carbon reductions. \autoref{fig:pareto-front-accuracy-carbon} shows, \systemName resides on the Pareto frontier, signifying its optimal balance between accuracy and carbon relative to baselines.

\vspace{0.1cm}
\noindent {\bf Key Point:} \emph{\emph{\systemName's} client selection policy (SteadyStep), integrated with CLP and a carbon-aware straggler mitigation technique, outperforms all baseline selection policies in terms of carbon cost across all \noniid levels. It matches or surpasses them in accuracy, while maintaining consistent training times.}

\vspace{-0.4cm}
\subsection{\systemName on Real-World Datasets}
\label{sec:generalization}

In the \autoref{sec:cost-aware} to \autoref{sec:policy-variants}, we present the evaluation results for the EMNIST dataset. Here, we evaluate \systemName with two additional, distinct ML datasets to demonstrate its generalizability. We use the Shakespeare and the Google Speech dataset, both are inherently \noniid, i.e., data heterogeneous. We compare \systemName's selection policy (\emph{SteadyStep}) to random and utility-based selection.

\begin{figure}[t]
    \centering
    \begin{tabular}{@{}c@{\hspace{-3mm}}c@{}}
    \includegraphics[width=0.5\linewidth]{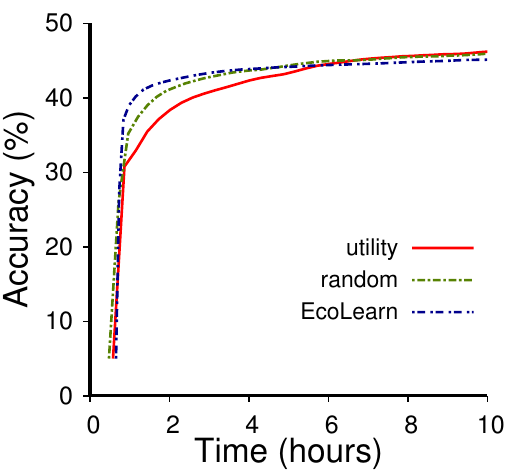} & 
    \includegraphics[width=0.5\linewidth]{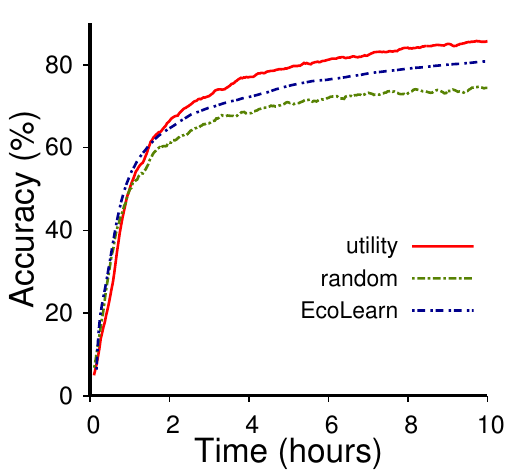}  \\
    (a) Shakespeare & (b) Google Speech
    \end{tabular}
    \vspace{-0.3cm}
    \caption{\textbf{\emph{Accuracy comparison of \systemName vs. random and utility-based baselines on Shakespeare and Speech datasets.}}}   
    \vspace{-0.4cm}
    \label{fig:shakespeare-speech-accuracies}
\end{figure}

\noindent
\textbf{Shakespeare.}
 \autoref{fig:shakespeare-speech-accuracies}(a) illustrates the accuracy of baseline selection policies and \systemName's selection policy on the y-axis against wall-clock time (hours) on the x-axis. We observe similar accuracy across all three selection strategies after training for the same duration, with utility-based selection performing slightly better. This behavior aligns with expectations and findings from recent studies \cite{wiesner2024fedzero, shin2022fedbalancer}. \autoref{tab:comparison} shows that \systemName achieves a \textbf{\emph{76.5\%}} and \textbf{\emph{73.1\%}} reduction in carbon cost compared to random selection and utility-based selection when trained for the same duration.

\noindent
\textbf{Google Speech.}
 \autoref{fig:shakespeare-speech-accuracies}(b) shows the accuracy of baseline selection policies (random- and utility-based) and \systemName's selection policy on the y-axis against wall-clock time (hours) on the x-axis. When trained for the same duration, \systemName achieves an accuracy of \emph{80.7\%}, and utility-based selection achieves \emph{85.8\%}. \systemName performs better the random baseline which achieves \emph{74.5\%} accuracy, as shown in \autoref{tab:comparison}. \autoref{tab:comparison} also demonstrates that \systemName reduces carbon cost significantly, surpassing both random  and utility-based selection with reductions up to \textbf{\emph{65.7\%}} and \textbf{\emph{68\%}}, respectively.

\vspace{0.1cm}
\noindent {\bf Key Point:} \emph{\systemName outperforms the carbon-agnostic baselines across different ML models and datasets suggesting that its carbon cost and accuracy tradeoffs are generalizable.}

\begin{table}[t] 
\centering
\footnotesize
\caption{{\bf \emph{Different selection policies across datasets.}}}
\label{tab:comparison}
\vspace{-0.3cm}
\begin{tabular}{|c|c|c|c|}
\hline
\textbf{Dataset} & \textbf{\begin{tabular}[c]{@{}c@{}}Selection\\Policies\end{tabular}} & \textbf{\begin{tabular}[c]{@{}c@{}}Convergence\\Accuracy\end{tabular}} & \textbf{\begin{tabular}[c]{@{}c@{}}Carbon\\Emissions (\%)\end{tabular}} \\
\hline
\hline
 & utility based  & 46.2 & 87.6 \\
\cline{2-4}
Shakespeare & Random & 45.9 & 100 \\
\cline{2-4}
 & EcoLearn & 45.1 & \textbf{\textit{23.5}} \\
\hline
\hline
 & utility based  & 85.8 & 107.3 \\
\cline{2-4}
Google Speech & Random & 74.5 & 100  \\
\cline{2-4}
 & EcoLearn & 80.7 & \textit{\textbf{34.3}}  \\
\cline{2-4}
\hline
\end{tabular}
\vspace{0.1cm}
\end{table}

\vspace{-0.1cm}
\section{Related Work}
\label{sec:related}
\noindent\textbf{Data distribution.}\hspace{2pt} FL uses privacy-preserving distributed algorithms~\cite{geyer2017differentially, pichai2019privacy}, keeping data on clients' premises, which may be geographically diverse~\cite{yang2018applied}. Unlike traditional ML, which assumes i.i.d data~\cite{jordan2015machine}, FL handles data from disparate probability distributions. We explore FL's non-i.i.d challenges~\cite{hsieh2020noniid} in relationship with carbon, utility, and CLP.

\noindent\textbf{Client selection strategies~\cite{client_selection_survey1, client_selection_survey2}.}\hspace{2pt} In FL, the challenge is building a robust model from clients while minimizing computation and communication. Prior work enhances models~\cite{caldas2018expanding}, optimizes communication~\cite{federated-learning-paper, wang2019adaptive, clp_accordion, li2020federated}, and reduces computations~\cite{xu2019elfish,caldas2018expanding}, but identifying high-value clients is under-explored. Some use statistical utility~\cite{oort-osdi, johnson2018training, katharopoulos2018not} or reduce clients per round while maintaining convergence~\cite{chen2022optimal, chen2018lag}. FLAME~\cite{cho2022flame} simulates client energy, but carbon cost and varying client selection strategies remain unaddressed.

\noindent\textbf{Carbon-aware client selection}\hspace{2pt}
Recent work~\cite{yousefpour2023green, qiu2023first, bian2024cafe} presents a study of carbon footprint of FL, presenting a solution to balance training performance and carbon emissions within a fixed budget across geo-distributed data centers whereas our approach dynamically adjusts the number of selected clients to reduce carbon emissions while improving performance. FedZero~\cite{wiesner2024fedzero} uses renewable resources to reduce carbon cost (orthogonal to \systemName), as it has different constraints, specifically, it selects clients with zero-carbon energy, which may result in some clients be unavailable for long periods extending the training time. 

\noindent\textbf{Critical learning period}\hspace{2pt} Prior work shows, CLP shapes the DNN model quality, where insufficient data during CLP causes permanent degradation~\cite{jastrzebski2019relation, golatkar2019time}. Early phases in FL training critically impact accuracy~\cite{clp_fl, clp_accordion} and are vulnerable to attacks~\cite{yan2023critical}. \systemName uses CLP insights to optimize scaling rates for better performance.

\section{Conclusion}
\label{sec:conclusion}
We introduce \systemName, which leverages clients' varying carbon cost.  \systemName selects clients based on their carbon cost and statistical utility to reduce carbon emissions, while maintaining accuracy.  Additionally, our investigation into critical learning periods inspired dynamic client selection, and scaling of clients to optimize cost without a significant impact on accuracy. To further reduce carbon, we optimize the tradeoff between wasted carbon due to stragglers and reduced training time due to straggler mitigation techniques. Our evaluation shows that \systemName reduces the carbon footprint of FL training (by up to $10.8$$\times$) while maintaining model accuracy and training time (within $\sim$$1$\%) compared to state-of-the-art approaches. 

\balance
\bibliographystyle{plain}
\bibliography{paper}

\end{document}